\documentclass[conference]{IEEEtran}
\usepackage{graphics} 
\usepackage{epsfig} 
\usepackage{mathptmx} 
\usepackage{amsmath} 
\usepackage{amssymb}  
\usepackage{graphicx}
\usepackage{eucal}
\usepackage{todonotes}[inline]  

\usepackage{float}
\usepackage[ruled,vlined]{algorithm2e}
\usepackage{pgfplotstable}
\usepackage{pgfplots}
\usepackage{tikz}
\usetikzlibrary[pgfplots.groupplots]
\usepgfplotslibrary{fillbetween}
\pgfplotsset{compat=1.5}
\usepackage{subcaption}
\usepackage{caption}
\usepackage{xcolor}
\usepackage{enumerate}
\usepackage{makecell}
\usepackage{colortbl}

\usepackage{graphicx}
\usepackage{tabularx,ragged2e,booktabs}
\usepackage{url}
\usepackage{siunitx}
\usepackage{rotating}
\usepackage{amssymb}
\usepackage{tabularx}
\usepackage{pgfplots}
\usepackage{csvsimple}
\usepackage{multirow}
\usepackage{cite}
\usepackage{float}
\usepackage{graphicx}
\usepackage{booktabs}

\ifCLASSINFOpdf
\else
\fi
\DeclareMathOperator*{\argmax}{arg\,max}

\begin{document}
%
\title{An Empirical Evaluation of Various Information Gain Criteria for Active Tactile Action Selection for Pose Estimation}



%
\author{\IEEEauthorblockN{Prajval Kumar Murali\IEEEauthorrefmark{1}\IEEEauthorrefmark{2},
Ravinder Dahiya\IEEEauthorrefmark{2}, and
Mohsen Kaboli\IEEEauthorrefmark{1}\IEEEauthorrefmark{3}}
\IEEEauthorblockA{\IEEEauthorrefmark{1}BMW Group, M\"unchen, Germany, \IEEEauthorrefmark{2}University of Glasgow, Scotland, \IEEEauthorrefmark{3}Radboud University, Netherlands. \\Email: prajval-kumar.murali@bmwgroup.com, ravinder.dahiya@glasgow.ac.uk, mohsen.kaboli@bmwgroup.com}}

\maketitle
\begin{abstract}
Accurate object pose estimation using multi-modal perception such as visual and tactile sensing have been used for autonomous robotic manipulators in literature. Due to variation in density of visual and tactile data, we previously proposed a novel probabilistic Bayesian filter-based approach termed translation-invariant Quaternion filter (TIQF) for pose estimation. As tactile data collection is time consuming, active tactile data collection is preferred by reasoning over multiple potential actions for maximal expected information gain. 
In this paper, we empirically evaluate various information gain criteria for action selection in the context of object pose estimation. We demonstrate the adaptability and effectiveness of our proposed TIQF pose estimation approach with various information gain criteria. We find similar performance in terms of pose accuracy with sparse measurements across all the selected criteria. 
\end{abstract}
\begin{IEEEkeywords}
Pose estimation, active action selection, information theory, tactile perception
\end{IEEEkeywords}

%
\IEEEpeerreviewmaketitle

\section{Introduction}
\label{sec:introduction}
Pose estimation of objects is a critical component of the perception pipeline for robots to interact with the real world. Visual perception techniques have undergone tremendous progress in the recent times due to the advent of deep learning methodologies coupled with inexpensive cameras and depth sensors. However, visual perception is sensitive to environmental conditions, object properties and scene configuration and tactile sensing has been used in prior works which provides robust and high fidelity grounded measurements about the objects~\cite{Qiang-TRO-2020, dahiya2019large, dahiya2019skin, murali2021intelligent, dahiya2013robotic, murali2022deepdomain}. However, collection of large amounts of tactile data is time consuming, cumbersome and leads to sensor wear-and-tear. Hence intelligent data collection strategies are required for efficient tactile-based perception~\cite{murali2021active, murali2022active,kaboli2017tactile, kaboli2018active, kaboli2018robust, kaboli2019tactile, kaboli2016tactilehuman}.

While tactile data can be gathered in a random fashion or by a human-teleoperator, active touch tactics are favored since they allow for autonomous data collection and reduce duplicate data collection.
Uncertainty reduction using information theoretic metrics such as Shannon entropy~\cite{kaboli2019tactile}, Kullback–Leibler (KL) divergence~\cite{petrovskaya2016active}, mutual information~\cite{kaboli2017tactile} are typically used for action selection.
There are various strategies studied in literature for next best view selection which can also be extended to the tactile domain. Delmerico et al.~\cite{delmerico2018comparison} studied various volumetric information gain metrics for next best view selection. Zhang et al.~\cite{zhang2019beyond} devised an active strategy based on Fisher Information for active visual localisation. Similarly, Carillo et al.~\cite{carrillo2018autonomous} devised a novel strategy based on Shannon entropy and R\'enyi entropy for autonomous robot exploration. In the tactile domain, various strategies have been used for object property estimation, texture recognition, pose estimation and so on. Fishel and Loeb~\cite{fishel2012bayesian} implemented an active texture recognition strategy based on tactile sensing using the Bhattacharyya coefficient. Kaboli et al.~\cite{kaboli2019tactile} used the variance in the Gaussian Process model to select the next exploratory action. Similarly, Shannon entropy has been used for selecting actions that provides the maximum discriminatory information for object classification~\cite{kaboli2018active, feng2018active}. In our previous works~\cite{murali2021active, murali2022active} KL-divergence has been used for tactile action selection for object pose estimation. 

\textbf{Contribution: }In this article, we empirically evaluate various information theoretic criteria for selection of the next best action in the context of tactile-based localisation. We use our novel probabilistic translation-invariant Quaternion filter (TIQF) for pose estimation proposed in~\cite{murali2022active}. 
We empirically evaluate the following criteria: (a) Kullback-Liebler divergence, (b) R\'enyi divergence, (c) Wasserstein distance, (d) Fisher information metric and (e) Bhattacharya distance for computation of next best touch in simulation.

\section{Methodology}
\label{sec:methods}
\subsection{Translation-Invariant Quaternion Filter (TIQF)}
\label{sec:tiqf}
To solve the point cloud registration problem for pose estimation, we proposed our linear translation-invariant Quaternion filter (TIQF) in our previous work~\cite{murali2021active, murali2022active}. 
Given correspondences, the point cloud registration problem can be defined as:
\begin{equation}
     \mathbf{s}_i = \mathbf{R}\mathbf{o}_i + \mathbf{t} \quad i = 1, \dots N \quad ,
     \label{eq:generativemodel}
 \end{equation}
 where $\mathbf{s}_i \in \mathcal{S}$ denotes the scene point cloud and $\mathbf{o}_i \in \mathcal{O}$ denotes the model point cloud.
 The rotation $\mathbf{R} \in SO(3)$ and translation $\mathbf{t} \in \mathbb{R}^3$ are undefined and computed through point cloud registration by aligning $\mathbf{o}_i$ with $\mathbf{s}_i$. 
 We decouple the estimation of rotation and translation by computing the vectors between consecutive points:
\begin{align}
    \mathbf{s}_j - \mathbf{s}_i &= (\mathbf{R}\mathbf{o}_j + \mathbf{t}) - (\mathbf{R}\mathbf{o}_i + \mathbf{t}) \quad ,\\
    \mathbf{s}_{ji} &= \mathbf{R}\mathbf{o}_{ji}  \quad . 
    \label{eq:trans_invariance}
\end{align}

We cast the problem of estimating rotation to a Bayesian estimation framework and estimate it using a Kalman filter.
We define the state $\mathbf{x}$ of our filter as the rotation to be estimated. We can rewrite \eqref{eq:trans_invariance} as: 
\begin{equation}
    \widetilde{\mathbf{s}}_{ji} = \mathbf{x} \odot \widetilde{\mathbf{o}}_{ji} \odot \mathbf{x}^{*} \quad . 
    \label{eq:quat_objective}
\end{equation}
\begin{align}
    \begin{bmatrix}
        0 & -(\mathbf{s}_{ji} - \mathbf{o}_{ij})^T \\
        (\mathbf{s}_{ji} - \mathbf{o}_{ji}) & (\mathbf{s}_j + \mathbf{s}_i + \mathbf{o}_j + \mathbf{o}_i)^{\times}
        \end{bmatrix}_{4 \times 4} \mathbf{x} &= \mathbf{0} 
        \label{eq:expected_measurement}
\end{align}
Similar to~\cite{arun2019registration}, a \textit{pseudo measurement model} for the Kalman filter can be defined as:
\begin{align}
    \mathbf{H}_t \mathbf{x} &= \mathbf{z}^h \quad .
    \label{eq:measurement_model}
\end{align}
We assume that $\mathbf{x}$ and $\mathbf{z}_t$ are Gaussian distributed and the Kalman filter is defined as:
\begin{align}
    \mathbf{x}_{t} &= \bar{\mathbf{x}}_{t-1} - \mathbf{K}_t \left( \mathbf{H}_t \bar{\mathbf{x}}_{t-1} \right) \\
    \Sigma^{\mathbf{x}}_{t} &= \left( \mathbf{I} - \mathbf{K}_t \mathbf{H}_t \right) \bar{\Sigma}^{\mathbf{x}}_{t-1} \\
    \mathbf{K}_t &= \bar{\Sigma}^\mathbf{x}_{t-1} \mathbf{H}_t^T \left( \mathbf{H}_t\bar{\Sigma}^\mathbf{x}_{t-1} \mathbf{H}_t^T + \Sigma_t^{\mathbf{h}}\right)^{-1} \quad , \label{eq:kalman_equations}
\end{align}
where $\bar{\mathbf{x}}_{t-1}$ is the normalized mean of the state estimate at $t-1$, $\mathbf{K}_t$ is the Kalman gain and $\bar{\Sigma}^{\mathbf{x}}_{t-1}$ is the covariance matrix of the state at $t-1$. The parameter $\Sigma_t^{\mathbf{h}}$ is the measurement uncertainty that is state-dependent and defined as~\cite{choukroun2006novel}:
\begin{align}
    \Sigma_t^{\mathbf{h}} = \frac{1}{4}\rho\left[ tr(\bar{\mathbf{x}}_{t-1}\bar{\mathbf{x}}_{t-1}^T + \bar{\Sigma}^{x}_{t-1})\mathbb{I}_4 - (\bar{\mathbf{x}}_{t-1}\bar{\mathbf{x}}_{t-1}^T + \bar{\Sigma}^{x}_{t-1} )\right] \quad ,
    \label{eq:choukron}
\end{align}
where $\rho$ is a constant that is tuned empirically.
As the Kalman filter does not ensure the constraint on the state to represent a rotation, we normalize the state and associated uncertainty after a prediction step as
\begin{equation}
    \bar{\mathbf{x}}_{t} = \frac{\mathbf{x}_{t}}{||\mathbf{x}_{t}||_2} \quad \bar{\Sigma}^{\mathbf{x}}_{t} = \frac{\Sigma^{\mathbf{x}}_{t}}{||\mathbf{x}_{t}||_2^2} \quad .
\end{equation}
With each iteration of the Kalman filter, a rotation estimate is obtained and the translation is computed using Equation~\eqref{eq:generativemodel}. The process is repeated until the change in the rotation and translation between iterations is less than a specified threshold.

\subsection{Next Best Touch Selection}
\label{sec:active_selection}

\begin{table*}[t!]
\caption{Divergence/ distance measures for multivariate Gaussian distributions $p_i = \mathcal{N}(\mu_i, \Sigma_i)$ and $p_j = \mathcal{N}(\mu_j, \Sigma_j)$}
\label{tab:eq}
\centering
\resizebox{0.85\textwidth}{!}{%
\begin{tabular}{@{}lll@{}}
\toprule 
\multicolumn{1}{c}{Name} & \multicolumn{1}{c}{\textbf{$D(p_{i}||p_j)$}} & \multicolumn{1}{c}{Comments} \\ \midrule
Kullback-Leibler divergence                                    &   $\frac{1}{2}[ log\frac{|({\Sigma}_{j})|}{|({{\Sigma}}_{i})|} + tr({\Sigma}_{j}^{-1} {{\Sigma}}_{i})) - d + ({{\mu}_{i}} - {\mu}_{j})' {\Sigma}_{j}^{-1} ({{\mu}}_{i} - {\mu}_{j})]$ &               d = 4 in our case                                                       \\
R\'enyi divergence                                          &         $ \frac{\alpha}{2}(\mu_i - \mu_j)'(\Sigma_\alpha)^*(\mu_i - \mu_j) - \frac{1}{2(\alpha-1)}log \frac{|(\Sigma_\alpha)^*|}{|\Sigma_i^{1-\alpha}||\Sigma_j^{\alpha}|} $
& $(\Sigma_{\alpha})^* = \alpha\Sigma_j + (1-\alpha)\Sigma_i$
\\
Fisher Information metric                                        &        $|\Sigma_j^{-1}(\mu_i - \mu_j)|^2 + tr(\Sigma_j^{-2}\Sigma_i -2\Sigma_j^{-1} + \Sigma_i^{-1})$                                                                 &\\
Bhattacharya distance                                     &       $\frac{1}{8}(\mu_i - \mu_j)'\Sigma^{-1}(\mu_i -\mu_j) + \frac{1}{2} log (\frac{|\Sigma|}{\sqrt{|\Sigma_i||\Sigma_j|}})$                                                                  & $\Sigma = \frac{\Sigma_i + \Sigma_j}{2}$\\
2-Wasserstein distance$^2$                                      &       $ |(\mu_i - \mu_j)|^2 + tr(\Sigma_i + \Sigma_j -2\sqrt{\sqrt{\Sigma_i}\Sigma_j\sqrt{\Sigma_i}}))$                                                              &    \\ \bottomrule
\end{tabular}
}
\end{table*}

As tactile data collection is time consuming, redundant data collection must be avoided. The next best touch action to perform is chosen as the action that reduces the overall uncertainty of the pose estimate.
A set of actions $\mathcal{A}$ are sampled uniformly on the faces of the bounding box on the current pose estimate of the object.
An action is defined as a ray represented by a tuple $\mathbf{a} = (\mathbf{n}, \mathbf{d})$, with $\mathbf{n}$ as the start point and $\mathbf{d}$ the direction of the ray. 
The optimal action $\mathbf{a}^{*}_{t}$ is chosen as the one that \textit{maximizes} the overall \textit{Information Gain}.
However, as the predicted measurements $\mathbf{z}_{t}$ are hypothetical, we can approximate our action-measurement model $p(\mathbf{z}_{t} | \mathbf{x}, \mathbf{a}_{t})$ as the ray-mesh intersection of the predicted action and the mesh of the model at the current estimated pose.
The hypothetical pose for each predicted action and predicted measurement is calculated using the TIQF algorithm as the \textit{one-step look ahead}. In order to calculate the information gain, we empirically compare a few well known information theoretic criteria as follows:
\begin{enumerate}
    \item \textit{Kullback–Leibler divergence (KL)}~\cite{kullback1951information} (or relative entropy) measures the how different one probability distribution is from another. For two discrete probability distributions $p_i$ and $p_j$ defined in the probability space $s\in \mathcal{S}$, $D_{KL}(p_i || p_j) = \sum_{s\in\mathcal{S}} p_i(s) log\frac{p_i(s)}{p_j(s)}$.  
    \item \textit{R\'enyi divergence}\cite{renyi1961measures} generalises the KL divergence and is defined as: $D_{\alpha}(p_i || p_j) = \frac{1}{1 - \alpha}log(\sum_{s\in\mathcal{S}}\frac{p_i^{\alpha}(s)}{p_j^{\alpha-1}(s)})$ for $0 < \alpha < \infty$ and $\alpha \neq 1$. For limiting case $\alpha \rightarrow 1$, the R\'enyi divergence is the same as KL divergence.
    \item \textit{Fisher information metric}\cite{fisher1922mathematical} measures the amount of information that an observable random variable $X$ carries about an unknown parameter $\theta$ upon which the probability of $X$ depends. It is defined as second derivative of the KL divergence.
    \item \textit{Bhattacharya distance}~\cite{bhattacharyya1943measure} measures the relative closeness of two probability distributions. It is defined as $D_B(p_i || p_j) = -\ln(\sum_{s\in\mathcal{S}}\sqrt{p_i(s)p_j(s)})$.
    \item \textit{Wasserstein distance}~\cite{olkin1982distance} is a way to compare two probability distributions, where one distribution is derived from the other by small, non-uniform perturbations (random or deterministic). It is defined as $W_p(\lambda,\nu) = (inf(\mathbb{E}[d(X,Y)^p])^{1/p}$ for two distributions $\lambda, \nu$ and the infimum is taken over all joint distributions of the random variables $X$ and $Y$ with marginals $\lambda$  and $\nu$  respectively~\cite{enwiki}.
\end{enumerate}
Given that the prior and posterior are multivariate Gaussian distributions, we have closed form solutions for each of the divergence or distance metrics as described in Table~\ref{tab:eq}. 
Therefore we perform the most optimal action $\mathbf{a}_{t}^*$ given by
\begin{align}
    &\mathbf{a}_{t}^* 
    &= \argmax_{\hat{\mathbf{a}}_{t}} D({p(\mathbf{x} | \hat{\mathbf{z}}_{1:t},  \hat{\mathbf{a}}_{1:t})}||{p(\mathbf{x} | \mathbf{z}_{1:t-1}, \mathbf{a}_{1:t-1})} \quad . \label{eq:kl_div_prio_post}
\end{align}

\section{Experimental Results}
\label{sec:experiment}
\begin{figure*}[t!]
    \centering
    \includegraphics[width = 0.9\textwidth, height = 5cm]{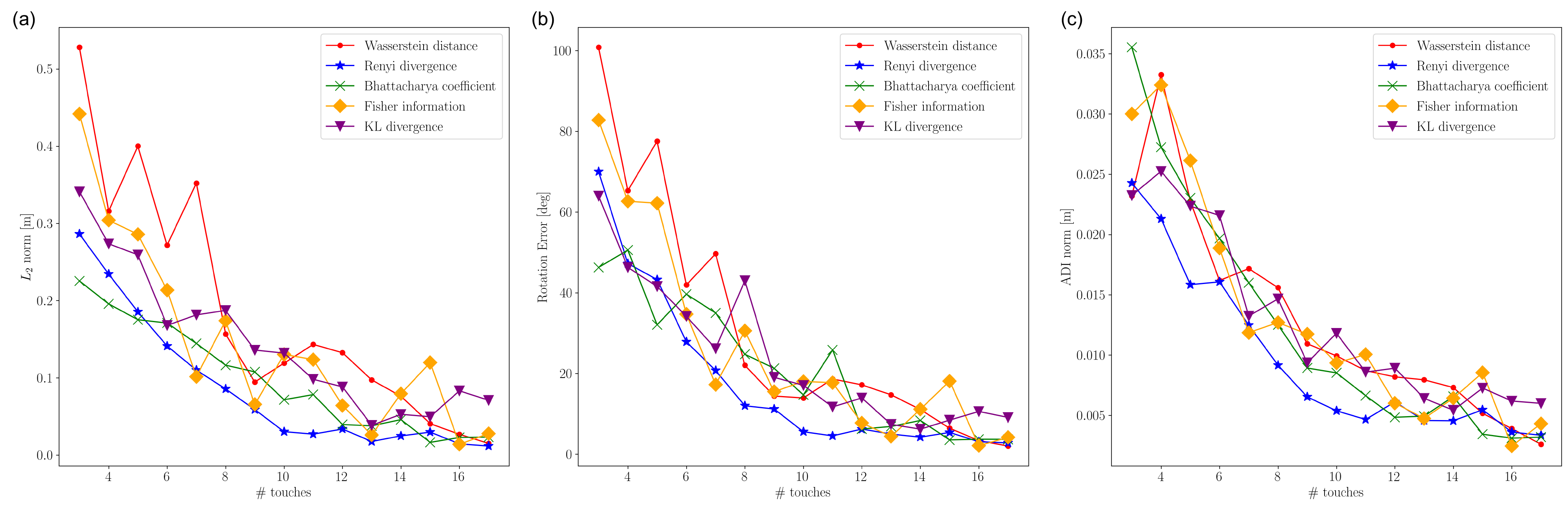}
    \caption{Simulation results on the Stanford Scanning Bunny dataset for the various information gain metrics: (a) Average $L_2$ norm of the position error with number of touch measurements, (b) average $L_2$ norm of the rotation error with number of touch measurements, (c) average ADI with number of touch measurements for 6 repeated runs.}
    \label{fig:plots}
\end{figure*}

\begin{figure*}[t!]
    \centering
    \includegraphics[width = \textwidth, height = 4cm]{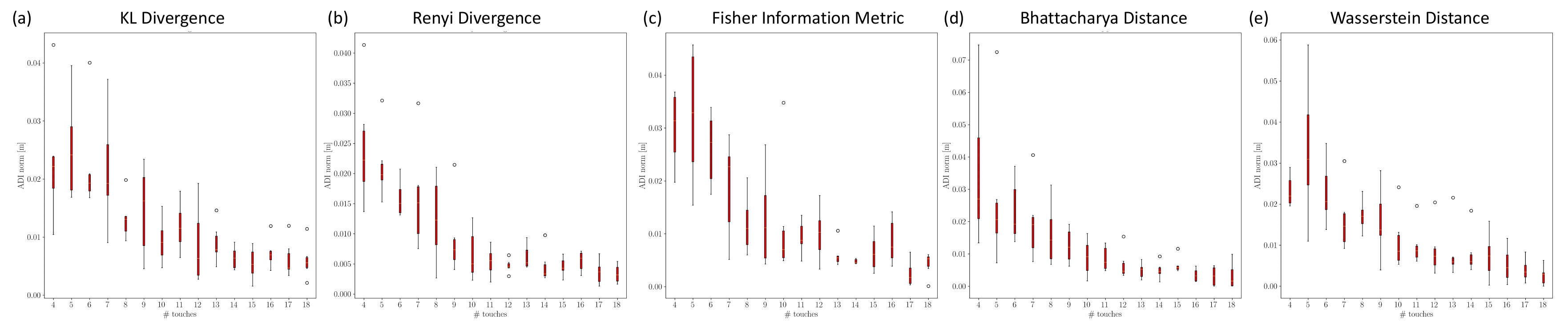}
    \caption{Box-and-whisker plots of the ADI metric for each criteria with increasing number of measurements.}
    \label{fig:plots_std}
\end{figure*}
We performed simulation experiments on the Bunny dataset of the Stanford Scanning Repository for calculating the next best touch using various information gain criteria. 
Noise was added that was randomly sampled from a normal distribution $\mathcal{N}(0, 5\times10^{-3})$ to the model cloud. The initial pose is randomnly sampled from $[-50, 50]mm$ and $[-30^o, 30^o]$ for translation and rotation respectively. The initial state $\mathbf{x}_0$ is set as the initial pose and the initial covariance $\Sigma^{\mathbf{x}}_0$ is set to $10^4*\mathbb{I}_4$.
To initialise the TIQF algorithm, a minimum number of measurements (3 points) are needed which are sampled by performing random touch actions. The active touch action selection is started from the 4th touch onwards.
In particular for R\'enyi divergence, we used an $\alpha =0.3$ that was empirically tuned.
All simulation experiments were executed on a workstation running Ubuntu 18.04 with 8 core Intel i7-8550U CPU @ 1.80GHz and 16 GB RAM.
We show the results of the simulation experiments for 6 repeated runs on each metric and show the average $L_2$ norm of the absolute error in position (m), rotation ($^{o}$) and Average Distance of model points with Indistinguishable views (ADI) metric~\cite{hinterstoisser2012model}. 

\subsection{Discussion}
The results of the experiments are presented in the Figure~\ref{fig:plots} and Figure~\ref{fig:plots_std}.
Across all the selected criteria, the pose error iteratively reduces with increasing number of measurements that are actively selected. In fact, for all the criteria the ADI metric is $<1$cm for $15$ measurements. Hence it shows that the active touch strategy using any information gain criteria with our proposed pose estimation approach helps to reduce the pose error with few measurements. We note very similar performance for each criteria and it is not straightforward to conclude if any particular criteria is better. We see comparatively lower variance for R\'enyi divergence and Wasserstein distance. In terms of accuracy, we note that KL divergence has comparatively slightly lower accuracy with other criteria however still within the acceptable accuracy range of $<1$cm.   
Hence, we provide initial empirical evaluation in simulation showing the adaptability of our proposed TIQF pose estimation approach with various information gain criteria. 

\section{Conclusion}
\label{sec:conclusions}
In this work, we empirically evaluated various information gain criteria for action selection in the context of object pose estimation in simulation. Our proposed TIQF algorithm for pose estimation allows for all criteria to have closed form solutions for the next best action with marginal computation time overhead. This work also provides the theoretical framework for employing various well known and uncommon information theoretic criteria for action selection. As future work, we will investigate the same criteria with real world data collected from novel tactile sensors~\cite{ozioko2021sensact, escobedo2020energy, ntagios2020robotic, mukherjee2021bioinspired}.

\bibliographystyle{IEEEtran}
\bibliography{IEEEabrv,root}
\end{document}